\documentclass[10pt,twocolumn,letterpaper]{article}
\usepackage{wacv}
\usepackage{times}
\usepackage{epsfig}
\usepackage{graphicx}
\usepackage{amsmath}
\usepackage{amssymb}
\usepackage{newtxmath}
\usepackage{multirow}

\usepackage[accsupp]{axessibility}
\wacvfinalcopy 

\ifwacvfinal
\usepackage[breaklinks=true,bookmarks=false]{hyperref}
\else
\usepackage[pagebackref=true,breaklinks=true,colorlinks,bookmarks=false]{hyperref}
\fi

\begin{document}

\title{Visual Understanding of Complex Table Structures from Document Images}

\author{Sachin Raja\\
IIIT-Hyderabad\\
{\tt\small sachin.raja@research.iiit.ac.in}
\and
Ajoy Mondal\\
IIIT-Hyderabad\\
{\tt\small ajoy.mondal@iiit.ac.in}
\and
Jawahar C V\\
IIIT-Hyderabad\\
{\tt\small jawahar@iiit.ac.in}
}

\maketitle
\thispagestyle{empty}

\begin{abstract}
Table structure recognition is necessary for a comprehensive understanding of documents. Tables in unstructured business documents are tough to parse due to the high diversity of layouts, varying alignments of contents, and the presence of empty cells. The problem is particularly difficult because of challenges in identifying individual cells using visual or linguistic contexts or both. Accurate detection of table cells (including empty cells) simplifies structure extraction and hence, it becomes the prime focus of our work. We propose a novel object-detection-based deep model that captures the inherent alignments of cells within tables and is fine-tuned for fast optimization. Despite accurate detection of cells, recognizing structures for dense tables may still be challenging because of difficulties in capturing long-range row/column dependencies in presence of multi-row/column spanning cells. Therefore, we also aim to improve structure recognition by deducing a novel rectilinear graph-based formulation. From a semantics perspective, we highlight the significance of empty cells in a table. To take these cells into account, we suggest an enhancement to a popular evaluation criterion. Finally, we introduce a modestly sized evaluation dataset with an annotation style inspired by human cognition to encourage new approaches to the problem. Our framework improves the previous state-of-the-art performance by a 2.7\% average F1-score on benchmark datasets.
\end{abstract}

\section{Introduction} \label{section_introduction}

\begin{figure}[htp!]
\begin{center}
\fbox{\includegraphics[width=0.97\linewidth]{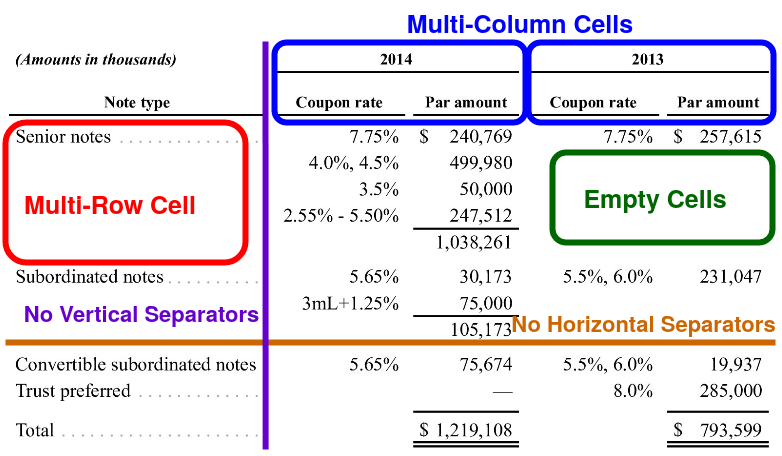}}
\end{center}
\caption{Demonstrates the challenges in table structure recognition task including absence of horizontal and vertical separators, multi-row/column spanning cells and empty cells. \label{fig_table_issues}}
\end{figure}

A fine-grained understanding of complex document objects such as tables, charts, and graphs in document images is challenging. We focus on table structure recognition, which is a precursor to semantic table understanding. Table structure recognition generates a machine-interpretable output for a given table image, which encodes its layout according to a pre-defined standard~\cite{table_splitting,li2019tablebank,paliwal2019tablenet,zhong2019image,chi2019complicated,xue2019res2tim,raja_eccv_2020}. Table structure recognition is difficult due to (a) inconsistency in size and density of tables, (b) absence of horizontal and/or vertical separator lines, (c) variation in table cells’ shapes and sizes, (d) table cells spanning multiple rows and (or) columns, (e) presence of empty cells, and (f) cells with multi-line content~\cite{hu1999medium,wang2004table,itonori1993table,green1995recognition,kieninger1998table,tupaj1996extracting}. Figure~\ref{fig_table_issues} visually illustrates some of the challenges.

Structure recognition of tables generally requires it to be broken down into cells first and then building associations between them. Cell detection is carried out using either visual or linguistic cues or both. As a precursor to obtain a good  structure recognition performance, it is imperative to detect cells that are highly accurate and  closely overlap with the ground truth. In few instances where access to machine-readable {\sc pdf}s is available, it becomes easier to identify content and its location for every table cell. Detection of table cells as independent objects is challenging, as discussed earlier. Contrarily, since tables generally adhere to an inherent structural alignment, it is relatively easier to locate columns and rows. However, that would split cells that span multiple rows/columns. In this work, we locate table cells independently and through detection of rows and columns while preserving the multi-row and multi-column spanning structures. Our results demonstrate improved F1-scores for cell detection and better localization of empty cells.

This brings up an interesting thought: ``How to interpret table cells without content and whether they carry any semantic meaning or not?". The absence of text in a table region
may or may not suggest the presence of empty table cells, which are therefore difficult to detect. In most cases, cells that have no content might carry implicit semantic meanings. For example, an empty cell in a numeric column in balance sheets would either indicate a zero value or `not applicable'. Similarly, the row header cell corresponding to the `total’ or `sum’ values is usually left blank. There might also be cases where an empty cell might span multiple rows and/or columns, such as a row header cell. In such instances, not correctly detecting empty cells would result in a loss of information during semantic parsing of the tables. Therefore, we emphasize the detection of empty cells and propose enhancing the existing vision-based criteria~\cite{icdar_2019} to consider empty cells for evaluation.

The natural follow up question becomes: ``What characterizes a good cell detection performance in a visual context?". In natural object detection, Intersection over Union (IoU) measure estimates of how well an object is detected. However, there are two concerning factors for cell detection: (i) How are the ground truth cell bounding-boxes annotated? (ii) What is the IoU threshold value used to compute evaluation metrics? For table cells, most datasets~\cite{qasim2019rethinking,chi2019complicated,deng2019challenges,li2019tablebank,zhong2019image,zheng2021global} have cell box annotation that spans the smallest rectangle encapsulating its content. This annotation style misses on the bounding boxes for empty cells and on cells' inherent alignment constraints. Further, most cell detection methods~\cite{icdar_2019,zheng2021global} evaluate using an Intersection over Union (IoU) threshold of $0.6$, which might not always correspond to capturing the entire cell content. In light of these challenges, we believe it is important for a cell detection method to perform well on high IoU thresholds. In that regard, there also arises a need for a standard evaluation dataset. Its ground truth cell boxes preserve their native alignment constraints (just as we humans perceive tables) and have annotations for empty cells. We present Table Understanding for Complex Documents ({\sc tucd}) as an evaluation dataset consisting of $4500$ manually annotated table images from business domain with a high diversity of table layouts having complex structures (samples shown in the supplementary material).

To detect table cells, we propose TOD-Net, where we augment the cell detection network of {\sc t}ab{\sc s}truct-{\sc n}et~\cite{raja_eccv_2020} with additional loss components to further improve the table object performance (rows/columns/cells) detection. These losses (formulated as regularizers) improve cell detection performance on high IoU thresholds by pairwise modelling of structural constraints. It allows for an improved bounding box detection despite presence of non-informative visual features in a specific table region using information from other cells detected in a different region of the table.

Once table cells are located precisely, extracting structure as an {\sc xml} or any other predefined format is relatively easier. However, for extremely dense tables with many multi-row and multi-column spanning cells, it may still be challenging to build associations between cells that are far apart in the two-dimensional space. To handle this problem, we propose {\sc tsr-n}et for structure recognition which uses the existing {\sc dgcnn} architecture~\cite{qasim2019rethinking}. Our formulation uses rectilinear adjacencies instead of row/column adjacencies~\cite{qasim2019rethinking,raja_eccv_2020}. Recursive parsing of rectilinear adjacencies helps to build better long-range visual row/column associations. 

\begin{figure*}[htp!]
\begin{center}
\fbox{\includegraphics[width=0.96\linewidth]{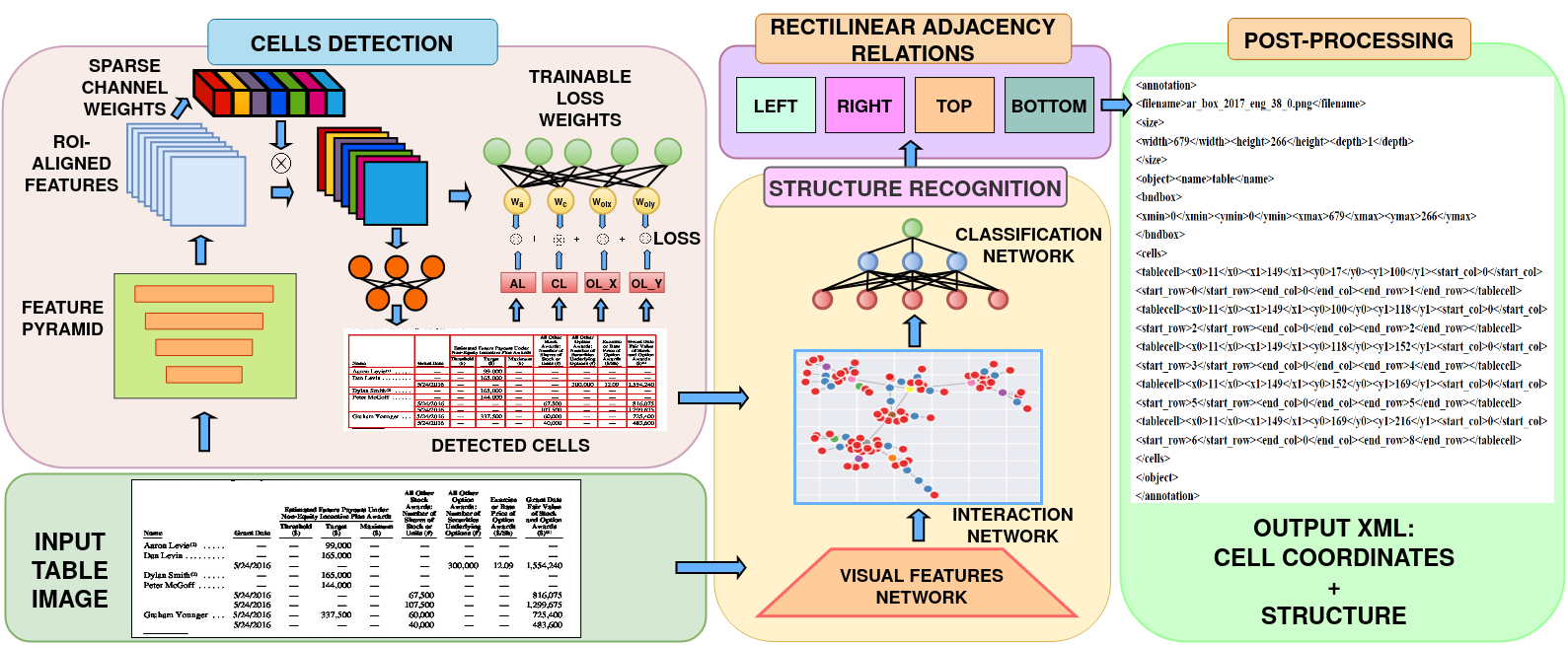}}
\end{center}
\caption{Shows our approach. Cell detection is done using {\sc tod-n}et. Bounding boxes used as an input by the structure recognition model (based on {\sc dgcnn}~\cite{qasim2019rethinking}, which predicts rectilinear adjacencies. These are then collectively used by the post-processing step to generate output {\sc xml} containing structure).\label{fig_tod_net}}
\end{figure*}

Our contributions can be summarized as follows:
\begin{itemize} 

\item Introduce channel attention~\cite{noori2019attention} for table object detection and define two additional regularizers --- continuity and overlapping loss between every pair of cells in addition to the alignment loss from~\cite{raja_eccv_2020}. We use trainable loss-weights for these losses and formulate a min-max optimization problem for faster convergence.

\item Formulate structure recognition using rectilinear adjacencies instead of row/column adjacencies, eliminating the need for complex post-processing heuristics for generating row and column spanning information for every cell.

\item Introduce modestly sized manually annotated {\sc tucd} as an evaluation dataset comprising $4500$ table images from publicly available annual reports. 

\item Suggest improvements to the existing criterion proposed in~\cite{icdar_2019} for a stricter evaluation of table structure recognition and demonstrate significantly improved performance on relatively higher IoU thresholds of $0.7$ and above compared to the state-of-the-art methods.

\item We demonstrate improved performance on cell detection through intermediate row and column detection tasks. 
\end{itemize} 

\section{Related Work} \label{section_related_work}

Early methods~\cite{wang2004table,hu1999medium} on table structure recognition primarily depend on hand-crafted features and heuristics (horizontal and vertical ruling lines, spacing, and geometric analysis). However, these usually make strong assumptions about table layouts for a domain agnostic algorithm. Some recent data-driven methods include works by ~\cite{adiga2019table,van2020table,prasad2020cascadetabnet,wang2021tablelab}.
Cognitive methods in this space broadly classified into five categories --- image-to-sequence models~\cite{li2019tablebank,bao2018table,Khan_2019}, segmentation networks~\cite{schreiber2017deepdesrt,nishida2017understanding,paliwal2019tablenet,qiao2021lgpma}, graph formulations~\cite{qasim2019rethinking,chi2019complicated,raja_eccv_2020}, conditional generative adversarial networks~\cite{le2019extracting} and a recent multi-modal method by ~\cite{zhang2021split}. A combination of heuristics and deep learning methods was also proposed~\cite{table_splitting} based on splitting the table into sub-cells, and then merging semantically connected sub-cells to preserve the complete table structure. These algorithms are robust to input types (scanned images or native digital) and do not generally make assumptions about the layouts. They are data-driven, and easy to fine-tune across different domains. Some methods that use linguistic context were proposed by \cite{bao2018table,nishida2017understanding,deng2019table2vec}. Many invoice-specific table extraction models have also been proposed~\cite{table_invoice,holevcek2019line}.

Recently, many researchers have opted for a graph-based formulation of the problem as a graph is inherently an ideal data structure to model associations between entities~\cite{qasim2019rethinking,chi2019complicated,raja_eccv_2020}. Raja~\cite{raja_eccv_2020} proposed a first end-to-end object detection and graph based model for collective cells detection and structure recognition. Another recent work, {\sc gte-c}ell~\cite{zheng2021global}, follows a nested approach by first classifying whether a table includes ruling lines or not, and then uses specifically tailored heuristics to identify the table structure. While these methods contribute to significant progress, they make certain assumptions like the availability of accurate word bounding boxes, machine readable {\sc pdf} documents, and others, as additional inputs~\cite{nishida2017understanding,qasim2019rethinking,chi2019complicated,table_splitting}. Contrarily, the {\sc t}ab{\sc s}truct-{\sc n}et~\cite{raja_eccv_2020} does not make any such assumptions and produces adjacency relations and cell locality information as the output. However, it fails to capture empty cells accurately and, in many cases, results in a significant overlap between detected cells. Further, its structure recognition module failed to correctly identify row/column associations between far-apart cells in case of dense tables.

Given the recent successes in natural object detection~\cite{Cai_2018_CVPR,tan2020efficientdet,Xu_2019_ICCV}, and the effectiveness of attention in improving its performance~\cite{Wang_2020_CVPR,Wang_2020_CVPR,wang2019towards,wu2020exploring}, we base our cell detection model on the object detection paradigm.
Our work aims to localize low-level table objects better on higher IoU thresholds, including empty cells. Our work also improves long range associations for structure recognition through rectilinear adjacency based formulation.

\section{Proposed Method} \label{proposed_method}

We formulate the table understanding problem at two levels --- \textit{low-level}, i.e., detection of table objects (rows, columns, and cells) and \textit{high-level}, i.e., physical structure recognition by building associations between cells. Most existing methods define table cells as the smallest polygon that encapsulates its content. This has two shortcomings. (i) It misses on the alignment and continuity constraints that are very natural to human cognition; and (ii) it misses on empty cells that usually carry important semantic meanings. Inspired by human cognition, we say that table cells, in addition to completely encapsulating their content, should adhere to alignment~\cite{raja_eccv_2020}, continuity and non-overlapping constraints, which in-turn makes it easier to locate table columns and rows as independent objects.

As discussed in Section~\ref{section_introduction}, many existing methods assume pre-located cell content and target only high-level structure understanding. Usually, table cells' coordinates are obtained by parsing corresponding {\sc pdf}/{\sc l}a{\sc t}e{\sc x} documents which may not always be available. Several methods also use {\sc ocr} tools to extract cell contents, resulting in the loss of intra-cell associations and structural alignment. Further, the absence of cell content makes it difficult to consider empty cells for structure recognition. In many real-world documents, empty cells carry a semantic meaning and must be associated with the table to obtain an accurate table structure. Not taking them into account might lead to false negatives and, in-turn, incorrect structure~\cite{raja_eccv_2020}. To localize table cells, we experiment by solving cell detection directly and through the intersection of predicted rows and columns. After locating all cells, we build rectilinear associations between every pair by formulating the problem as a graph. 

Our solution progresses in three steps, as shown in Figure~\ref{fig_tod_net}, --- (i) table cell detection using visual cues, (ii) structure recognition by forming rectilinear associations through a graph-based formulation, and (iii) collating bounding boxes and rectilinear associations to obtain row and column spanning values for every table cell. 

\subsection{Cell Detection}\label{cell_detection}

We aim to detect table cells in two ways --- (i) by locating them as independent objects and (ii) by first locating rows and columns as independent objects and then using intersections to obtain cell coordinates. We target row, column, and cell detection as object detection tasks using our Table Object Detection Network ({\sc tod-n}et shown in Figure~\ref{fig_tod_net}), built on top of the cell detection network of {\sc t}ab{\sc s}truct-{\sc n}et~\cite{raja_eccv_2020, he2017mask}. Our augmentations to the existing architecture aim to model the constraints associated with table objects to ensure adjacent cells' continuity and non-overlap. We use sparse channel weights on the {\sc roi} aligned feature maps to predict table objects' bounding boxes (cells, rows, and columns). We also formulate the problem as min-max optimization using adaptable loss weights for the three structural regularizers --- alignment loss~\cite{raja_eccv_2020}, continuity loss, and overlap loss. 

\paragraph{\textbf{Notations:}}

Let $\vmathbb{X}$ denote the set of table images;  $SR(i)$, $SC(i)$, $ER(i)$, and $EC(i)$ represent start-row, start-column, end-row, and end-column indices respectively; and $x_{1i}$, $y_{1i}$, $x_{2i}$ and $y_{2i}$ represent bounding box coordinates start-x, start-y, end-x, and end-y, respectively of the object $i$. $i$ and $j$ denote two table objects (row/column/cell). $L_{m}$ denotes the sum of {\sc rpn} class loss, {\sc rpn} bounding box regressor loss, Mask {\sc r-cnn} class loss, Mask {\sc r-cnn} bounding box regressor loss, and mask loss. $L_{al}$, $L_{cl}$, $L_{ol}^{x}$, and $L_{ol}^{y}$ represent alignment loss, continuity loss, and overlap losses along X and Y directions respectively; and $W_{al}$, $W_{cl}$, $W_{ol}^{x}$, and $W_{ol}^{y}$ represent corresponding learnable weights.

\paragraph{\textbf{Continuity Loss:}} 

The intuition behind adding continuity loss is that horizontally adjacent objects should end and start at the same x-coordinate and vertically adjacent objects end and start at the same y-coordinate. Continuity loss is given in Eq.~(\ref{eqn_continuity_loss})
\begin{equation}
\begin{aligned}
L_{cl}^{row} &= \sum_{i, j} ||y_{1i}-y_{2j}||_2^2 \cdot \  \vmathbb{I}{\big (SR(i) == ER(j) + 1\big )} \\ 
L_{cl}^{col} &= \sum_{i, j} ||x_{1i}-x_{2j}||_2^2 \ \cdot  \vmathbb{I}{\big (SC(i) == EC(j) + 1\big )} \\
L_{cl} &= L_{cl}^{row} + L_{cl}^{col}. \label{eqn_continuity_loss} 
\end{aligned}
\end{equation}
This loss helps to predict well-aligned coordinates by accurately capturing the background or non-text region associated with objects that are significantly wider or longer than the text region contained in them. 

\paragraph{\textbf{Overlapping Loss:}} 

We introduce overlapping loss as an L2 regularizer to minimize overlapping regions between every pair of predicted table objects. During the calculation, the overlap of an object with itself does not account for the loss. Further, it is computed independently along X and Y directions (as given in Eq.~\ref{eqn_overlap}).
\begin{equation}
\begin{aligned}
L_{ol}^{x} = \sum_{i, j} ||(min(x_{2i}, x_{2j}) - max(x_{1i}, x_{1j})||_2^2 \cdot  \vmathbb{I}\big (i != j\big ) , \\
L_{ol}^{y} = \sum_{i, j} ||(min(y_{2i}, y_{2j}) - max(y_{1i}, y_{1j}))||_2^2 \cdot  \vmathbb{I}\big (i!= j\big ) \label{eqn_overlap}
\end{aligned}
\end{equation}

\paragraph{\textbf{Trainable Loss Weights:}} 

We incorporate trainable loss weights for four different structure components as regularizers (alignment, continuity, and overlap loss along X and Y directions) for every region of interest ({\sc roi}) independently such that the weights add up to one. This allows for a dynamic emphasis on different structural constraints for different {\sc roi}s based on their visual characteristics during training. We model the optimization problem as a min-max optimization problem as follows:

\begin{equation}
\begin{aligned}
\vmathbb{L}(\vmathbb{X}, \theta_{m}, \theta_{W}) = min_{\theta_{m}} \big(\ L_{m}(\theta_{m})\big)\ +\max_{\theta_{W}}\big( \\ 
W_{al}(\theta_{W}) \cdot L_{al}(\theta_{m})\ +\  W_{cl}(\theta_{W}) \cdot L_{cl}(\theta_{m})\ +\ \\ 
W_{ol}^{x}(\theta_{W}) \cdot L_{ol}^{x}(\theta_{m})\ +\ 
W_{ol}^{x}(\theta_{W}) \cdot L_{ol}^{y}(\theta_{m})\ \big) \\ 
\ni W_{al}\ +\ W_{cl}\ +\ W_{ol}^{x}\ +\ \ W_{ol}^{y} &=\ 1 \label{eqn_final}
\end{aligned}
\end{equation}

Since we need to minimize the objective loss (as given in Eq.~(\ref{eqn_final})) over $\theta_{m}$ and maximize over $\theta_{W}$, the parameter updates are given by the following Eq.~(\ref{eqn_weight_update})
\begin{equation}
\begin{aligned}
\theta^{t+1}_{m} &= \theta^{t}_{m}\ -\ \eta\ \cdot\  \nabla_{\theta^{t}_{m}} \big (\vmathbb{L}(\vmathbb{X}, \theta^{t}_{m}, \theta^{t}_{W}) \big ) \\
\theta^{t+1}_{W} &= \theta^{t}_{W}\ +\ \eta\ \cdot\  \nabla_{\theta^{t}_{W}} \big (\vmathbb{L}(\vmathbb{X}, \theta^{t}_{m}, \theta^{t}_{W}) \big ),
\label{eqn_weight_update}
\end{aligned}    
\end{equation}
where $\eta$ is the learning rate. 
Formulation based on a min-max optimization problem using trainable loss weights (by allowing for weighting different regularizers differently based on RoI’s visual features) not only improves optimization speed, but also proves useful during post-processing. We use the predicted values of loss weights during the test time to identify and correct overlapping or misaligned cells. Our experiments suggest that high overlapping loss weights were observed during test time for dense table images. Similarly, high alignment values and continuity losses were observed for multi-column or multi-row spanning header cells where the text was not aligned in the center.

\paragraph{\textbf{Channel Attention:}}

To detect table objects' start and end coordinates, specific visual patterns such as separator lines or non-text regions need to be present. These visual patterns differ significantly from general object detection problems where different shaped edges and textures are essential to distinguish different types of objects. For the detection of table objects, the distinguishing visual clues occur in particular regions of every {\sc roi}. In order to localize table cells, specific set of visual features contribute. For example, a column (or a row) would start or end at an x (or y)-coordinate where around that region, either a vertical (or a horizontal) separator or non-text/background is observed along the length (or width) of the image. This motivates us to incorporate L1-regularized channel-wise attention to look for specific sparse patterns to detect cell bounding boxes accurately. The attention-mechanism we use is based on the architecture proposed by~\cite{noori2019attention} and is shown in Figure~\ref{fig_tod_net}.

\subsection{Structure Recognition}\label{tsr_net} 

We formulate the table structure recognition as a graph learning problem similar to~\cite{qasim2019rethinking}. However, instead of creating row and column adjacency matrices, we create four rectilinear matrices such as left ($M_{l}$), right ($M_{r}$), top ($M_{t}$), and bottom ($M_{b}$) $\in R^{n\times n}$, where n denotes the number of detected cells. For the top rectilinear matrix, $M_{t}$, the element at $M_{t}(i,j)$ indicates whether cell $j$ is at the top of cell $i$. Similarly, we create left $M_{l}$, right $M_{r}$, and bottom $M_{b}$ matrices. Formulating the problem in this way allows for better capturing of long-range dependencies for dense tables particularly. We use four instances of the {\sc dgcnn} architecture proposed in~\cite{qasim2019rethinking} to predict the four rectilinear matrices. The {\sc dgcnn} consists of three components --- (i) a visual network to generate a visual feature map corresponding to the input table image, (ii) an interaction network to capture associations between cells from the visual features and coordinates of table cells, and (iii) a classification network to determine if a pair of table cells are left/right/top/bottom adjacent. The training happens in two steps. In the first step, we use ground-truth boxes, and in the second step, we fine-tune the models using predictions of {\sc tod-n}et on the training dataset. The training adjacencies are obtained by identifying the largest overlapping ground truth cell corresponding to the prediction.

\subsection{Post-processing}

Firstly, we fine-tune the predicted cell bounding boxes using Tesseract's~\cite{smith2007overview} word bounding boxes to ensure that the predicted cell boundary region does not pass through any text region. Once cell bounding boxes and rectilinear adjacency matrices are obtained, the next step is to figure out row and column spanning values for every cell. The maximum count of left and right adjacencies is obtained recursively to obtain row span for cell $i$. Similarly, to obtain column span for cell $i$, the maximum count of top and bottom adjacencies is obtained recursively. Finally, start-row ($SR$), end-row ($ER$), start-column ($SC$), and end-column ($EC$) indices for every cell are obtained by sorting the coordinates based on start-x and end-x coordinates along with the row and column spans obtained using the rectilinear adjacency matrices. The use of rectilinear adjacencies accounted for reduced use of heuristics and improved F1 scores for structure recognition. Our final output comes out as an {\sc xml} that contains bounding boxes along with the row and column spans for every cell given a table image.

\begin{table}[htp!]
\addtolength{\tabcolsep}{-3.5pt}
\begin{center}
\begin{tabular}{|l|l|r|r|r|} \hline
 &\textbf{Document} &\textbf{Alignment} &\textbf{\#Train} &\textbf{\#Test} \\
\textbf{Dataset} &\textbf{Domain} &\textbf{Constraint}  &\textbf{Image} &\textbf{Image}  \\ \hline\hline
{\sc icdar-2013} &Business &$\times$   &-    &156 \\
{\sc unlv}       &Business &\checkmark   &-    &558 \\
c{\sc td}a{\sc r}    &Business &$\times$  &600 &150 \\
{\sc s}ci{\sc tsr} &Scientific &$\times$ &12{\sc k} &3{\sc k} \\
{\sc t}able2{\sc l}atex  &Scientific &$\times$ &447{\sc k} &9{\sc k} \\   
{\sc t}able{\sc b}ank &Scientific &$\times$ &145{\sc k} &1{\sc k} \\
{\sc p}ub{\sc T}ab{\sc n}et &Scientific &$\times$ &420{\sc k} &40{\sc k} \\
{\sc f}in{\sc t}ab{\sc n}et &Business &$\times$  &91{\sc k}    &10{\sc k} \\
{\sc tucd} (our) &Business &\checkmark &- &4.5{\sc k} \\ \hline
\end{tabular}
\end{center}
\caption{Presents statistics of datasets for table structure recognition. Only {\sc t}able{\sc b}ank~\cite{li2019tablebank} is dedicated for logical table structure recognition. All other datasets are used for physical table structure recognition.   \label{table_statistic_dataset}}
\end{table}

\subsection{Datasets}\label{tucd_dataset}

Most datasets~\cite{qasim2019rethinking,chi2019complicated,deng2019challenges,li2019tablebank,zhong2019image,zheng2021global} use words or cell content as low-level entities to build inter-tabular relationships. Similarly, there exist inconsistencies in the datasets for predicting the physical or logical structure of tables. This presents a fundamental challenge to evaluate and compare various methods for table structure recognition directly. ~\cite{qasim2019rethinking,chi2019complicated,li2019tablebank,zhong2019image} introduced many large-scale automatically generated datasets, but they do not accurately represent real-world complex tables as seen in the business documents~\cite{zheng2021global,shahab2010open,gobel2013icdar}. Another matter of concern is the style of annotation. As humans, we think of tables adhering to specific structural and alignment constraints --- (i) cells belonging to the same row should start and end at the same start-y and end-y coordinates respectively, (ii) cells belonging to the same column should start and end at the same start-x and end-x coordinates respectively, (iii) cells starting at column $i$ should have the same start-x coordinate as the end-x coordinate of column $i-1$, (iv) cells starting at row $i$ should have the same start-y coordinate as the end-y coordinate of row $i-1$, (v) no overlap between any pair of table cells.
Presently, {\sc unlv}~\cite{shahab2010open} is the only dataset where ground-truth preserves this inherent structural alignment between cells. However, this dataset is limited in size, language, and domain variations for evaluating a deep learning-based method. Other datasets~\cite{chi2019complicated,zhong2019image,zheng2021global,gobel2013icdar} have annotations such that a cell's bounding box is the smallest rectangle that encapsulates its content. This leads to non-annotation for empty cells and loss of alignment between cells in the same and adjacent rows/columns.

{\sc tucd} dataset is dedicated to evaluation of cells detection and structure recognition for business documents. It consists of $4500$ table images collected from the publicly available annual reports in English and non-English languages (e.g., French, Japanese, Russian, and others) of more than ten years from twenty-nine different companies\footnote{TUCD dataset is available at https://github.com/sachinraja13/TUCD}. The ground truth {\sc xml} for a table image contains the coordinates of bounding boxes of cells and their row and column spans. Table~\ref{table_statistic_dataset} lists the statistics of different structure recognition datasets available for training and testing.

\begin{table*}[ht!]
\addtolength{\tabcolsep}{-3.35pt}
\begin{center}
\begin{tabular}{|l|l|l|l|l|l|l|l|l|l|l|l|l|l|l|l|l|l|} \hline
 & & &\multicolumn{15}{|c|} {\textbf{Average Over Test Set}} \\ 
\cline{4-18}
 &\textbf{Training} & &\multicolumn{3}{|c|}{\textbf{ICDAR-2013}} &\multicolumn{3}{|c|}{\textbf{SciTSR}}&\multicolumn{3}{|c|}{\textbf{SciTSR Comp}} &\multicolumn{3}{|c|}{\textbf{ICDAR-19}} & \multicolumn{3}{|c|}{\textbf{TUCD}}  \\
 \cline{4-18}
\textbf{Method} &\textbf{Dataset} &\textbf{EC} &\textbf{P}$\uparrow$ &\textbf{R}$\uparrow$ &\textbf{F1}$\uparrow$ &\textbf{P}$\uparrow$ &\textbf{R}$\uparrow$ &\textbf{F1}$\uparrow$ &\textbf{P}$\uparrow$ &\textbf{R}$\uparrow$ &\textbf{F1}$\uparrow$ &\textbf{P}$\uparrow$ &\textbf{R}$\uparrow$ &\textbf{F1}$\uparrow$ &\textbf{P}$\uparrow$ &\textbf{R}$\uparrow$ &\textbf{F1}$\uparrow$ \\ \hline\hline
{\sc d}eep{\sc d}e{\sc srt}~\cite{schreiber2017deepdesrt} &{\sc icdar-13} & &\textit{0.96} &\textit{0.87} &\textit{0.91}	&- &- &- &-	&- &-&-&-&-&- &- &- \\
{\sc splerge}(H)~\cite{table_splitting} &Private & &\textit{0.96} &\textit{0.95} &\textit{0.95} &- &- &-&- &- &- &- &- &-&- &- &- \\
{\sc split}~\cite{table_splitting} &Private  &{\sc sec} &\textit{0.87} &\textit{0.87} &\textit{0.87} &0.92	&0.97 &0.97 & 0.91 & 0.88 & 0.90 &0.70 &0.67 &0.69 &0.87 &0.86 &0.86\\
{\sc t}ab{\sc s}truct-{\sc n}et~\cite{raja_eccv_2020} &{\sc s}ci{\sc tsr}&  &\textit{0.92} &\textit{0.90} &\textit{0.91}	&\textit{0.93} &\textit{0.91} &\textit{0.92}	&\textit{0.91} &\textit{0.88} &\textit{0.90}&\textit{0.60} &\textit{0.57} &\textit{0.58}	&0.90	&0.89	&0.90 \\
{\sc gte-c}ell~\cite{zheng2021global} &{\sc f}in{\sc t}ab{\sc n}et &  &\textit{0.96} &\textit{0.97} &\textit{0.96} &-	&- &- &-&- &-&-&- &- &-&- &- \\
{\sc sem}~\cite{zhang2021split} &{\sc s}ci{\sc tsr} &  &-&-&-&\textit{0.98} &\textit{0.97} &\textit{0.97}&\textit{0.97} &\textit{0.95} &\textit{0.96} &-&- &-&-&- &- \\
{\sc lgpma}~\cite{qiao2021lgpma} &{\sc s}ci{\sc tsr} &  &\textit{0.93} &\textit{0.98} &\textit{0.95} &\textbf{\textit{0.98}} &\textbf{\textit{0.99}} &\textbf{\textit{0.99}} &\textbf{\textit{0.97}} &\textbf{\textit{0.99}} &\textbf{\textit{0.98}}&-&- &-&-&- &- \\\hline 
{\sc d}eep{\sc d}e{\sc srt}~\cite{schreiber2017deepdesrt} &{\sc f}in{\sc t}ab{\sc n}et & &0.82 &0.80 &0.81	&0.87 &0.85 &0.86 &0.85 &0.83 &0.84 & 0.55 & 0.51 &0.53 &0.73	&0.70 &0.72 \\
{\sc dgcnn}$^{\dagger}$~\cite{qasim2019rethinking,raja_eccv_2020}  &{\sc f}in{\sc t}ab{\sc n}et & &0.94 &0.93 &0.94 &0.91	&0.89 &0.90 &0.89	&0.88 &0.89 & 0.73 & 0.70 & 0.71 &0.89 &0.87	&0.88 \\
{\sc dgcnn}$^{\ddagger}$~\cite{qasim2019rethinking,raja_eccv_2020}  &{\sc f}in{\sc t}ab{\sc n}et & &0.96 &0.95	&0.96 &0.91	&0.90 &0.91 &0.90	&0.89 &0.89 & 0.76 & 0.73 & 0.74 &0.92 &0.91 &0.91 \\
{\sc t}ab{\sc s}truct-{\sc n}et~\cite{raja_eccv_2020} &{\sc f}in{\sc t}ab{\sc n}et  &{\sc sec}  &0.95 &0.94 &0.95 &0.90	&0.89	&0.90 &0.88	&0.87	&0.87 & 0.76 & 0.73 & 0.75 &0.91 &0.90 &0.90 \\
Ours$^{\dagger}$  &{\sc f}in{\sc t}ab{\sc n}et & &0.95 &0.95 &0.95 &0.92	&0.91 &0.92 &0.92	&0.90 &0.91 & 0.72 & 0.70 & 0.71 & 0.91 &0.90	&0.91 \\ 
Ours$^{\ddagger}$   &{\sc f}in{\sc t}ab{\sc n}et & &\textbf{0.98} &\textbf{0.97} &\textbf{0.97} &0.94 &0.92 &0.93 &0.93 &0.89& 0.91&\textbf{0.77}	&\textbf{0.76} &\textbf{0.77} &\textbf{0.94}	&\textbf{0.93} &\textbf{0.93}\\ \hline \hline
{\sc d}eep{\sc d}e{\sc srt}~\cite{schreiber2017deepdesrt} &{\sc f}in{\sc t}ab{\sc n}et & &0.74 &0.71 &0.73	&0.82 &0.80 &0.81& 0.80 &0.79 &0.79 & 0.53 & 0.48  &0.50 &0.70	&0.68 &0.69 \\
{\sc split}~\cite{table_splitting} &Private & &0.83 &0.81	&0.82 &0.89 &0.87	&0.88&0.87 &0.87	&0.87 & 0.68 & 0.66 &0.67 &0.82 &0.81 &0.81\\
{\sc dgcnn}$^{\dagger}$~\cite{qasim2019rethinking,raja_eccv_2020} &{\sc f}in{\sc t}ab{\sc n}et  & &0.87 &0.85 &0.86 &0.89	&0.87 &0.88 &0.87	&0.85 &0.86 &0.69 &0.67 &0.68 &0.86 &0.85	&0.85 \\ 
{\sc dgcnn}$^{\ddagger}$~\cite{qasim2019rethinking,raja_eccv_2020}   &{\sc f}in{\sc t}ab{\sc n}et  & &0.90 &0.89 &0.89 &0.88	&0.85 &0.86 &0.86	&0.84 &0.85 &0.71 &0.69 &0.70 &0.89 &0.88 &0.89 \\
{\sc t}ab{\sc s}truct-{\sc n}et~\cite{raja_eccv_2020} &{\sc s}ci{\sc tsr}  & {\sc nec}  &0.89 &0.87 &0.88	&0.90	&0.87	&0.88&0.88	&0.86	&0.87 &0.54 & 0.49 & 0.51	&0.84	&0.83	&0.83 \\
{\sc t}ab{\sc s}truct-{\sc n}et~\cite{raja_eccv_2020} &{\sc f}in{\sc t}ab{\sc n}et &&0.90 &0.87 &0.89	&0.88	&0.85	&0.86&0.86	&0.84	&0.85 &0.70	&0.69 &0.70 &0.88 &0.86 &0.87 \\
Ours$^{\dagger}$   &{\sc f}in{\sc t}ab{\sc n}et~ & &0.91 &0.90 &0.90 &0.90	&0.86	&0.88&0.88	&0.84	&0.86 &0.70 &0.67 &0.68	&0.90	&0.88 &0.89 \\
Ours$^{\ddagger}$ &{\sc f}in{\sc t}ab{\sc n}et & &\textbf{0.93} &\textbf{0.92} &\textbf{0.92} &\textbf{0.91}	&\textbf{0.88} &\textbf{0.89}&\textbf{0.89}	&\textbf{0.87} &\textbf{0.88} &\textbf{0.73} & \textbf{0.72} & \textbf{0.72} &\textbf{0.92} &\textbf{0.91}	&\textbf{0.92} \\
\hline
\end{tabular}
\end{center}
\caption{Compares various methods for table structure recognition on {\sc icdar-2013}, {\sc sci-tsr}, {\sc sci-tsr comp}, {\sc icdar-19} and {\sc tucd} datasets. Scores in italics are directly reported from corresponding papers. For others, we use open source implementations and pre-trained models released by authors. For {\sc d}eep{\sc d}e{\sc srt}~\cite{schreiber2017deepdesrt}, we use our implementation.  {\sc ec:} indicates evaluation criteria, {\sc sec}: indicates standard evaluation criteria, and {\sc nec}: indicates new evaluation criteria. P: indicates precision, R: indicates recall, and F1: indicates F1 score. {\sc tod-n}et$^{\dagger}$: indicates {\sc tod-n}et for direct cell detection and {\sc tod-n}et$^{\ddagger}$: indicates cell detection using intersection of {\sc tod-n}et results row and column predictions, {\sc dgcnn}$^{\dagger}$ indicates {\sc tod-n}et$^{\dagger}$+{\sc dgcnn}+{\sc pp}, {\sc dgcnn}$^{\ddagger}$ indicates {\sc tod-n}et$^{\ddagger}$+{\sc dgcnn}+{\sc pp} {\sc ts}-{\sc n}et indicates {\sc t}ab{\sc s}truct-{\sc n}et, Ours$^{\dagger}$ indicates {\sc tod-n}et$^{\dagger}$+{\sc tsr}+{\sc pp}, Ours$^{\ddagger}$ indicates {\sc tod-n}et$^{\ddagger}$+{\sc tsr}+{\sc pp} and (H) indicates dataset specific heuristics. For comparison on {\sc icdar-2013} using {\sc sec}, {\sc icdar-2013} text-based evaluation was used. All other results are based on a fixed IoU threshold of $0.6$. For the {\sc nec}, we additionally consider empty cells for evaluation. \label{table_comparison_result}}
\end{table*}

\subsection{Training and Evaluation}\label{gt_pre} 

We use {\sc f}in{\sc t}ab{\sc n}et~\cite{zheng2021global} dataset to train {\sc tod-n}et for cell, row, and column detection. Since {\sc f}in{\sc t}ab{\sc n}et has bounding boxes wrapped around the cell's content, we pre-process the ground truth to obtain cell level coordinates (refer supplementary paper)\footnote{Please refer supplementary material for dataset preprocessing, postprocessing, implementation and additional quantitative and qualitative results.}.  The resulting dataset follows all the constraints that we model in the {\sc tod-n}et. For evaluation also, we pre-process {\sc icdar-2013}~\cite{gobel2013icdar}, c{\sc td}a{\sc r}~\cite{icdar_2019}, {\sc s}ci{\sc tsr}~\cite{chi2019complicated}, {\sc p}ub{\sc t}ab{\sc n}et~\cite{zhong2019image} and {\sc f}in{\sc t}ab{\sc n}et~\cite{zheng2021global} datasets before computing IoU with the corresponding predictions. Since {\sc unlv}~\cite{shahab2010open} and {\sc tucd} datasets already have annotations for cells adhering to alignment constraints, we directly used them for evaluation. Further, during training and evaluation, we use the non-maximal suppression threshold of $0.8$ during proposal generation to reduce the false negatives substantially. We train {\sc tsr-n}et in two steps: In the first stage, we use pre-processed ground-truth cell boxes and corresponding start-row, start-column, end-row, and end-column indices to generate target rectilinear adjacency matrices. In the second stage, we generate predictions of the training set using {\sc tod-n}et to compute its overlap with the ground-truth to find start-row, start-column, end-row, and end-column indices for every predicted box. We accordingly generate target rectilinear adjacency matrices for training on the predicted boxes.

\begin{figure}[htp!]
\begin{center}
\includegraphics[width=0.9\linewidth, height=0.12\textwidth]{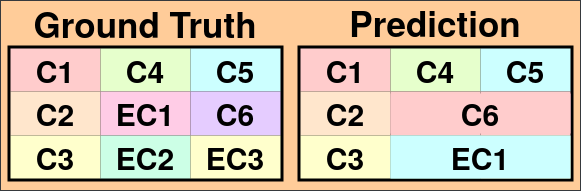}
\end{center}
\caption{Shows sample ground truth and predicted bounding boxes of cells for evaluation. Assume Cs to be cells with content and ECs to be cells without content. Also, assume detection of table cells merges EC1 and C6 in row $2$ and EC2 and EC3 in row $3$. Our proposed evaluation criteria additionally penalize (EC1, EC2) and (EC2, EC3) as false negatives. \label{fig_eva}}
\end{figure}

\subsection{Evaluation Protocol}

In literature, researchers~\cite{gobel2013icdar,shahab2010open,chi2019complicated} use precision, recall, and F1 scores to evaluate the performance of table's physical structure recognition. Adjacency relations for every true positive cell are generated with their horizontal and vertical neighbors to assess structure recognition performance. The predicted relation list is then compared with the ground truth list to calculate precision, recall, and F1 scores. However, these criteria do not consider empty cells that are not surrounded by non-empty cells to calculate performance scores. Since most existing methods use pre-located table cells as inputs, this does not cause any problem. However, as a result of cell detection, these empty cells might get merged with neighboring cells containing content or false positives, disturbing the overall table structure (as shown in Figure~\ref{fig_eva}). Henceforth, for the end-to-end structure recognition of given table images only, we suggest taking into account empty cells to calculate precision, recall, and F1 scores correctly. For table object (row, column, and cell) detection (both empty and with content), we calculate precision, recall, and F1 scores for an IoU threshold of $0.6$.

\section{Results}

This work presents a comprehensive analysis of results to understand the impact of architectural designs, modifications to the evaluation criteria, and optimization characteristics. For this purpose, we provide a four-fold analysis - comparative analysis with existing methods in the literature, analysis on varying IoU thresholds for cell detection, an ablation study showing the effectiveness of design choices and impact of loss weights on optimization speed.

\begin{table}[htp!]
\addtolength{\tabcolsep}{-5.0pt}
\begin{center}
\begin{tabular}{|l|l|l|l|l|l|} \hline
& &\textbf{FinTabNet} & \textbf{ICDAR-13} &\textbf{Sci-TSR} &\textbf{TUCD} \\ \cline{3-6}
\textbf{Method} &\textbf{IoU} &\textbf{TSR-F1}$\uparrow$ &\textbf{TSR-F1}$\uparrow$&\textbf{TSR-F1}$\uparrow$ & \textbf{TSR-F1}$\uparrow$  \\ \hline\hline
{\sc ts}-{\sc n}et  &	    &0.898 & 0.904 & 0.876 & 0.900  \\
Ours$^{\dagger}$  & 0.5 &0.906 & 0.903 & 0.880 & 0.889  \\
Ours$^{\ddagger}$ &     &\textbf{0.944} & \textbf{0.904} &  \textbf{0.894} & \textbf{0.918} \\\hline
{\sc ts}-{\sc n}et    & 	&0.848 & 0.886 &  0.864 & 0.871  \\
Ours$^{\dagger}$     &0.6  &0.892 & 0.903 & 0.878  & 0.889  \\
Ours$^{\ddagger}$ &  &\textbf{0.920}  & \textbf{0.904} &   \textbf{0.894} &  \textbf{0.918}\\\hline
{\sc ts}-{\sc n}et    & 	&0.704 & 0.720 &  0.682 & 0.722  \\
Ours$^{\dagger}$ &0.7  &0.802&  0.820 &  0.746 & 0.797  \\
Ours$^{\ddagger}$  & &\textbf{0.868}  & \textbf{0.852} & \textbf{0.823} & \textbf{0.839}\\\hline
{\sc ts}-{\sc n}et  & 	&0.496 & 0.597 & 0.565 & 0.582 \\
Ours$^{\dagger}$ &0.8  &0.561 &  0.675 &  0.637 &  0.659 \\
Ours$^{\ddagger}$ &    &\textbf{0.680} &  \textbf{0.748} & \textbf{0.714} & \textbf{0.735} \\\hline
{\sc ts}-{\sc n}et   & 	&0.120 & 0.292 & 0.255 & 0.289   \\
Ours$^{\dagger}$ &0.9  &0.325 & 0.307 & 0.296 & 0.301 \\
Ours$^{\ddagger}$   &     &\textbf{0.404}  &  \textbf{0.454} &  \textbf{0.368} & \textbf{0.408}\\\hline
\end{tabular}
\end{center}
\caption{Shows the comparison between the performances of the proposed network and {\sc t}ab{\sc s}truct-{\sc n}et ({\sc ts}-{\sc n}et)~\cite{raja_eccv_2020} on cell detection and table structure recognition of dataset over various IoU thresholds.{\sc tsr}: indicates table structure recognition. We use {\sc f}in{\sc t}ab{\sc n}et~\cite{zheng2021global} dataset for training.\label{table_iou_result_all}}
\end{table}

\paragraph{\textbf{Comparative Analysis}}

Table~\ref{table_comparison_result} shows results comparing our method against previously published on {\sc icdar-2013}, SciTSR, {\sc icdar-19} and {\sc tucd} datasets. Please note that in the first section of the table with evaluation using Standard Evaluation Criteria ({\sc sec}), we use {\sc icdar-2013} text-based measure for {\sc icdar-2013} dataset. On the contrary, corresponding {\sc sec}, we use IoU overlap based {\sc icdar-2019} evaluation criterion on SciTSR, {\sc icdar-19} and {\sc tucd} datasets. Further, for the second section of the table, that uses New Evaluation Criteria ({\sc nec}), we modify the IoU based {\sc icdar-2019} evaluation to additionally take into account adjacency relations between empty-empty and empty-non empty cells. For evaluating {\sc icdar-2013} dataset using {\sc nec}, we modify the ground truth to obtain cell-level boxes (as explained in Section~\ref{gt_pre}) and extend those to full rows and columns to obtain bounding box coordinates for empty cells (assuming no empty cells are multi-row/column spanning). Details of this step are provided in the supplementary section. For a fair comparison of our method against {\sc dgcnn}~\cite{qasim2019rethinking}, we use {\sc tod-n}et to obtain cell bounding boxes, obtain row and column adjacency matrices using {\sc dgcnn}~\cite{qasim2019rethinking} and use the open-source post-processing provided by~\cite{raja_eccv_2020}. In order to compare our method against others on {\sc tucd} dataset, we develop our implementation of {\sc d}eep{\sc d}e{\sc srt}~\cite{schreiber2017deepdesrt}, and use open source implementations of {\sc dgcnn} ({\sc ties})~\cite{qasim2019rethinking}, {\sc splerge}~\cite{table_splitting}, and {\sc t}ab{\sc s}truct-{\sc n}et~\cite{raja_eccv_2020}. For others, we directly report results from the corresponding papers. From the table, it is evident that formulating the problem using rectilinear adjacencies instead of row/column adjacency avoids errors in long visual ranges, relaxes heuristics in the post-processing method. Our method outperforms previous state-of-the-art on all three datasets by a reasonable difference of average F1-score on structure recognition. We further observe that empty cells account for an average of $12.3\%$ across {\sc unlv} and {\sc icdar-2013} datasets, where our method outperforms {\sc t}ab{\sc s}truct-{\sc n}et by $4.2\%$ F1 score.

Our solution however fails for very sparse tables where most of the cells are empty. We will add some qualitative examples in the supplementary material. Since rectilinear adjacencies are predicted between every pair of cells, inference time is in the order of square of number of cells located. For table images with 20 cells, inference time is about 10 seconds which goes upto 50 seconds for images with 200 cells.

\paragraph{\textbf{F1 based on Varying IoU Thresholds}}

For table cell detection, the IoU threshold becomes imperative as the penalty for loss of content or additional content detected from a localized table cell is high. Higher IoU also accounts for better structure recognition performance. Hence, a method's robustness can be established based on its performance under a higher IoU threshold. For this purpose, we evaluate the previously established benchmark~\cite{raja_eccv_2020} with our approach on IoU thresholds varying from $0.5$ up to $0.9$ as shown in Table~\ref{table_iou_result_all} according to our updated evaluation criteria that take into account empty cells present along the table extreme boundary regions. 

\paragraph{\textbf{Ablation Study}}

Table~\ref{table_loss_result} shows the ablation study of various enhancements to our {\sc tod-n}et. We observe that the addition of continuity loss improved the average F1 score by $0.8\%$. It especially proved helpful for table cells having a varying amount of text in table headers. For text consisting of large empty spaces with a very little text region, continuity loss helped detect the boxes that adhere to the inherent table alignment. We further observed that the addition of pairwise overlapping loss improved precision by $1.1\%$ and channel-wise multiplication of sparse channel weights further improved detection performance by $2.1\%$. Also, we observe that with the same weight initialization, the model with dynamic loss weights converges $15\%$ faster and slightly better by $0.4\%$.

\begin{table}[ht!]
\begin{center}
\begin{tabular}{|l|l|l|l|} \hline
 &\multicolumn{3}{|c|}{\textbf{Cell Detection}}  \\ \cline{2-4}
\textbf{Method} &\textbf{P}$\uparrow$ &\textbf{R}$\uparrow$ &\textbf{F1}$\uparrow$  \\ \hline\hline
Mask {\sc r-cnn}+{\sc al} &0.880 &0.862	&0.871  \\ \hline
Mask {\sc r-cnn}+{\sc al}+{\sc cl} &0.891 &0.868 &0.879 \\ \hline
Mask {\sc r-cnn}+{\sc al}+{\sc cl}+{\sc ol}  &0.907	&0.873 &0.890 \\ \hline
Mask {\sc r-cnn} +{\sc al}+ & & & \\
{\sc cl}+{\sc ol}+{\sc roi}\_Att.         &0.922 &0.900 &0.911 \\ \hline
Mask {\sc r-cnn}+{\sc al}+ & & & \\
{\sc cl}+{\sc ol}+{\sc roi}\_Att.+{\sc l}oss{\sc wt} &\textbf{0.926} &\textbf{0.904} &\textbf{0.915} \\ \hline
\end{tabular}
\end{center}
\caption{Shows the ablation study for cell detection on various structural constraints on baseline (Mask {\sc r-cnn}+{\sc al})~\cite{raja_eccv_2020}. We use new evaluation criteria with IoU threshold = $0.6$. {\sc tod:} indicates table object detection, {\sc al:} indicates alignment loss, {\sc cl:} indicates continuity loss, {\sc ol:} indicates overlapping loss, {\sc roi}\_Att.: indicates {\sc roi} attention, and {\sc l}oss{\sc wt}: indicates loss weights. We use {\sc f}in{\sc t}ab{\sc n}et~\cite{zheng2021global} dataset for training and evaluation. \label{table_loss_result}}
\end{table}

\section{Conclusion}

Our approach advances  both the formulation and the empirical performances compared to the state of the art methods. Major contributions include: (i) a formulation possibly closer to how human perceives tables (ii) architectural improvements to model problem-specific constraints, (iii) an adaptation of optimization, (iv) a novel {\sc tucd} dataset for evaluation and (iv) empirical evaluation extending the analysis to high IoU thresholds that improve practical usability.

Our work will advance the table understanding literature with immediate effect for better information extraction from business documents. We also believe, our insights in analyzing images with dense structured objects will impact wider categories of images captured in industrial vision setting, and crowded outdoor. Also, our dataset and improved evaluation can serve for a more robust evaluation of table structure. Further, the reasoning behind using trainable loss weights could be extended to niche domain specific problems (understanding of graphs/charts and establishing correct reading order from document images).

\section*{Acknowledgment}

This work is partly supported by MEITY, Government of India.

\end{document}